\documentclass{amsart}

\usepackage{amsmath, amssymb, amsthm, mathrsfs}
\usepackage{mathtools}
\usepackage[margin=1in]{geometry}

\def\utheta{\underline{\theta}}

\def\cC{{\mathcal C}}

\def\cL{{\mathcal L}}

\def\ux{{\underline{x}}}

\def\uy{{\underline{y}}}

\def\R{{\mathbb R}}

\def\eqn{\begin{eqnarray}}
\def\eeqn{\end{eqnarray}}
\def\eqnn{\begin{eqnarray*}}
\def\eeqnn{\end{eqnarray*}}

\def\prf{\begin{proof}}
\def\endprf{\end{proof}} 

\newcommand{\lra}{\longrightarrow}

\theoremstyle{definition}

\newtheorem{lemma}{Lemma}
\newtheorem{definition}{Definition}[section]

\newtheorem{theorem}{Theorem}[section]
\newtheorem{remark}{Remark}
\newtheorem{proposition}{Proposition}

\title[Zero loss in generic overparametrized DL]
{Zero loss guarantees and explicit minimizers for generic overparametrized Deep Learning networks}

\author{Thomas Chen}
\address[T. Chen]{Department of Mathematics, University of Texas at Austin, Austin TX 78712, USA}
\email{tc@math.utexas.edu} 
\author{Andrew G. Moore}
\address[A. G. Moore]{Department of Mathematics, University of Texas at Austin, Austin TX 78712, USA}
\email{agmoore@utexas.edu}

\begin{document}

\begin{abstract}
We determine sufficient conditions for overparametrized deep learning (DL) networks to guarantee the attainability of zero loss in the context of supervised learning, for the $\cL^2$ cost and {\em generic} training data. We present an explicit construction of the zero loss minimizers without invoking gradient descent. On the other hand, we point out that increase of depth can deteriorate the efficiency of cost minimization using a gradient descent algorithm by analyzing the conditions for rank loss of the training Jacobian. Our results clarify key aspects on the dichotomy between zero loss reachability in underparametrized versus overparametrized DL.  
\end{abstract}

\maketitle

\section{Introduction}

Sufficiently overparameterized deep feed-forward neural networks are capable of fitting arbitrary generic training data with zero $\cL^2$ loss. Later in this paper, we prove that this is true by an explicit construction of the zero loss minimizers without invoking gradient descent. On the other hand, analysis of this case shows something surprising: when the width is large enough, increasing the depth of the network can introduce obstructions to efficient cost minimization using gradient descent. This phenomenon is caused by rank loss of the training Jacobian, violating the assumptions of the simplicity guarantees in \cite{cheewa-5}. The rank of this object is very poorly studied, despite its relevance to the analysis of gradient descent trajectories. 

When the width is great enough, depth is not necessary at all to achieve zero loss, as linearly generic data may be fit by linear regression. However, this is an advantage from the point of view of qualitative analysis: by focusing on such a simple situation, the challenges to gradient flows introduced by depth become clearer by contrast. We prove sufficient conditions under various fairly general assumptions that generic training data ensures that the Jacobian has full rank for sufficiently overparameterized models, thus ensuring relative simplicity of gradient descent in these regions. Together with previous papers by one of the authors, our results shed light on key aspects of the dichotomy between zero loss reachability in underparametrized versus overparametrized DL. 

\section{Deep Neural Networks}

We define the DL network as follows.
The input space is given by $\R^M$, with training inputs $x_{j}^{(0)}\in \R^M$, $j=1,\dots,N$. 
The reference output vectors are given by a linearly independent family $\{y_{\ell}\in\R^Q | \ell=1,\dots,Q\}$ with $Q\leq M$, which label $Q$ classes. We introduce the map $\omega:\{1,\dots,N\}\rightarrow\{1,\dots,Q\}$, which assigns the output label $\omega(j)$ to the $j$-th input label, so that $x_{j}^{(0)}$ corresponds to  $y_{\omega(j)}$. 

%We define $\uy_\omega:=(y_{\omega(1)},\dots,y_{\omega(N)})^T\in\R^{NQ}$, where $A^T$ is the transpose of the matrix $A$. Let $N_i$ denote the number of training inputs belonging to the output vector $y_i$, $i=1,\dots,Q$. 

We assume that the DL network contains $L$ hidden layers. The $\ell$-th layer is defined on $\R^{M_\ell}$, and recursively determines
\eqn
	x_j^{(\ell)} = \sigma(W_\ell x_j^{(\ell-1)} + b_\ell) \;\;\in\R^{M_\ell}
\eeqn
with weight matrix $W_\ell\in\R^{M_\ell\times M_{\ell-1}}$, bias vector $b_\ell\in\R^{M_\ell}$. The map $\sigma:\R^{M\times N}$, $A=[a_{ij}]\mapsto [\sigma(a_{ij})]$ acts component-wise by way of the scalar activation function $\sigma:\R\rightarrow I\subseteq\R $ where $I$ is a connected interval. We assume that $\sigma$ has a Lipschitz continuous derivative, and that the output layer 
\eqn
	x_j^{(L+1)} = W_{L+1} x_j^{(L)} + b_{L+1} \;\;\in\R^{Q}
\eeqn
contains no activation function.

Let $\utheta \in\R^K$ enlist the components of all weights $W_\ell$ and biases $b_\ell$, $\ell=1,\dots,L+1$, including those in the output layer. Then,
\eqn
	K = \sum_{\ell=1}^{L+1} (M_\ell M_{\ell-1}+M_\ell )
\eeqn
where $M_0\equiv M$ for the input layer.

In the output layer, we denote  $x_j^{(L+1)}\in \R^Q$ by $x_j[\utheta]$ for brevity, and obtain the $\cL^2$ cost as
\eqn\label{eq-cC-def-1-0}
	\cC[\ux[\utheta]] &=&\frac1{2N}\big|\ux[\utheta]-\uy_\omega\big|_{\R^{QN}}^2
	\nonumber\\
	&=& \frac1{2N}\sum_j |x_j[\utheta]-y_{\omega(j)}|_{\R^Q}^2 \,,
\eeqn 
using the notation $\ux:=(x_1,\dots,x_N)^T\in\R^{QN}$, and $|\bullet|_{\R^n}$ for the Euclidean norm.

Training of the networks corresponds to finding the minimizer $\utheta_*\in\R^K$ of the cost, and we say that zero loss is achievable if there exists $\utheta$ so that $\cC[\ux[\utheta_*]] =0$. 

In matrix notation, we let
\eqn 
	X^{(\ell)} &:=& [\cdots x_{j}^{(\ell)} \cdots] \in\R^{M_\ell\times N}
	\nonumber\\
	Y_\omega &:=& [\cdots y_{\omega(j)} \cdots] \in\R^{Q\times N} \,.
\eeqn
Then, we have that
\eqn 
	X^{(\ell)} = \sigma(W_\ell X^{(\ell-1)}+B_\ell)
\eeqn 
where $B_\ell:=b_\ell u_N^T$ with $u_N:=(1,1,\dots,1)^T\in\R^N$.
Then, the solvability of 
\eqn 
	W_{L+1} X^{L} + B_{L+1} = Y_\omega 
\eeqn 
is equivalent to achieving zero loss.

\subsection{Underparametrized DL networks}

If $K < QN$, the DL network is underparametrized, and the map 
\eqn 
	f_{X^{(0)}}:\R^K\rightarrow R^{QN}
	\;\;\;,\;\;\;
	\utheta \mapsto \ux[\utheta]
\eeqn
is an embedding. Accordingly, for generic training data $X^{(0)}$ the zero loss minimizer of the cost is not contained in the submanifold $f_{X^{(0)}}(\R^K)\subset\R^{QN}$. 

However, if the training data is non-generic, zero loss can be achieved. As proven in \cite{cheewa-2,cheewa-4}, sufficient clustering and sufficient depth ($L\geq Q$) allows the explicit construction of global zero loss minimizers.

\subsection{Paradigmatic dichotomy}

In combination with Theorem \ref{thm:alt}, below, and results in subsequent sections of this paper, we establish the following paradigmatic dichotomy between underparametrized and overparametrized networks:
\\

\noindent
$\bullet$ {\em  
\underline{Underparametrized DL networks} with  $M\geq M_1\geq\cdots\geq M_L\geq Q$ layer dimensions, and (locally mollified) ReLU activation function
\begin{itemize}
\item[-]
cannot in general achieve zero loss for generic training data, 
\item[-]
but with sufficient depth, they are capable of achieving zero loss for non-generic training data with sufficient structure.
\item[-]
For sufficiently clustered or linearly sequentially separable training data, the zero loss minimizers are explicitly constructible without gradient descent, \cite{cheewa-2,cheewa-3}.
\end{itemize}

\noindent   
$\bullet$ \underline{Overparametrized networks} with layer dimensions $M=M_1=\cdots=M_L\geq Q$ and activations acting as diffeomorphisms $\sigma:\R^M\rightarrow\R_+^M$ 
\begin{itemize}
\item[-]
are capable of achieving zero loss if the training data are generic; the minimizers are explicitly constructible, without gradient descent.
\item[-]
However, increase of depth can decrease the efficiency of  cost minimization via gradient descent algorithm.
\item[-]
In deep networks with equal layer dimensions and certain regularity assumptions on the activation, zero loss minimizers can be explicitly constructed, but gradient descent might nevertheless fail to find them.
\end{itemize}
 
}

To this end, we prove the following theorem; a more complete discussion is given in later sections. The proof is constructive, i.e., it provides a method to obtain explicit zero loss minimizers when the hidden layers have equal dimensions.

\begin{theorem}\label{thm-overpar-expl-min-1-0}
Assume that $M=M_0>N$, and that all hidden layers have equal dimension, $M_\ell=M$, for all $\ell=1,\dots,L$, and that $Q\leq M$ in the output layer.

If we take any map $\sigma : \R^M \lra \R^M$ which is a local diffeomorphism at at least one point (this includes most common activation functions, including coordinate-wise hyperbolic tangent or ReLU), then, assuming generic training data, there exist choices of $W$ and $B$ at each layer such that the loss is zero.

%Secondly, if assume that $\sigma:\R\rightarrow\R_+$ is a diffeomorphism, so that $\sigma:\R^M\rightarrow\R_+^M$ also is a diffeomorphism, then for any fixed choices of $W_2, \dots W_{L+1}$ all invertible, there exist explicitly constructible choices of $W_1$ and $B_i$ for each $i$ such that there is zero loss.

\begin{proof}
    See Section \ref{sec:ffnn}.
\end{proof}
\end{theorem}

The proof given of Theorem \ref{thm-overpar-expl-min-1-0} chooses all weights and biases. In the particular case of a ReLU-like activation function, however, a more explicit construction is available using arbitrary fixed weights on most layers. We state this as an alternate version of the theorem:

\begin{theorem}
\label{thm:alt}
    Assume that $M=M_0>N$, and that all hidden layers have equal dimension, $M_\ell=M$, for all $\ell=1,\dots,L$, and that $Q\leq M$ in the output layer. Moreover, assume that $\sigma:\R\rightarrow\R_+$ is a diffeomorphism, so that $\sigma:\R^M\rightarrow\R_+^M$ also is a diffeomorphism. Assume $X$ is full rank and fixed. Pick any full rank matrices $W_2, \dots, W_L$. Then, there exist explicitly constructable choices of $W_1, W_{L+1}$, and $B_i$ for each $i$ such such that the loss is zero, and the choice of parameters is degenerate.
\begin{proof}
    See Section \ref{sec:appendix}.
\end{proof}
\end{theorem}

\begin{remark}
	We note that for the  gradient flow, $\nabla_{\utheta}\cC[\ux[\utheta]]$ includes powers of $\sigma'$ of degree up to $L$. Those derivatives can be arbitrarily small, for example, if $\sigma$ is taken to be a mollification of ReLU which is strictly monotone increasing, and $0<\sigma(x)<\epsilon\ll1$ for all $x<0$; in particular, $\sigma'(x)\searrow 0^+$ as $x\rightarrow-\infty$. 
	
	This implies that the convergence of gradient descent algorithms can be very slow along certain orbits, and that the numerical conditioning of the problem deteriorates with increase of the depth $L$.  
\end{remark}

\begin{remark}
	We demonstrated in the proof of Theorem \ref{thm:alt} (under the stated assumptions on the architecture of the DL network) that for generic training data, explicit zero loss minimizers can be straightforwardly constructed.
	
	However, for some orbits of the gradient flow, there could exist a finite time at which the right hand side of \eqref{eq-Xell-rec-expl-1-0}  fails to be positive and moves outside the domain of $\sigma^{-1}$. This will lead to a rank loss of the Jacobi matrix associated to $f_{X^{(0)}}:\R^K\rightarrow\R^{QN}$, and zero loss minimization might become unattainable, as shown in \cite{cheewa-5}; see the discussion in Section \ref{sec-overpar-1-0}.
\end{remark}

In the subsequent sections, we will prove more general and nuanced versions of this result.

For some thematically related background, see for instance \cite{hanrol,grokut-1,KMTM24,lcbh,manvanzde,nonreeste,PHD20} and the references therein.

\section{Notations}
 
We introduce the following notations, which are streamlined for our subsequent discussion.

\begin{definition}

The following notations will be used for the context of supervised learning.

%We need to start by laying out some symbology for the context of supervised learning which will be used constantly.

\begin{enumerate}
    \item Let $X \in \R^{M\times N}$ be the matrix of data. This is a matrix with $M$ rows and $N$ columns, consisting of $N$ data points, each of which is a vector in $\R^M$. The data  are represented by the columns of $X$, where the $i$th column is denoted by $X_i$. 
    \item Let $\Theta$ be a parameterized function realization map, considered as a map $\R^K \lra C^0(\R^M, \R)$, where $K$ is the number of parameters. We will use the notations $\Theta(\theta), f_\theta$, or $f$ interchangeably depending on the context. We can extend $f$ to a map $g: \R^{M \times N} \lra \R^N$ by defining $g^i(Y) = f(Y_i)$, where $Y$ is any data matrix.
    \item Define the Jacobian matrix $D$ as follows. Let the element of $D$ at the $i$th row and $j$th column be equal to $\partial g^i / \partial \theta^j$ (recall that $g$ depends on $\theta$ because it is defined in terms of $f = \Theta(\theta)$). It follows that $D \in \R^{N \times K}$.
    \item Let $y \in \R^N$ be the vector of \textit{labels}, i.e. the intended outputs for the datapoints.
    \item We define the \textit{loss} as a function of $\theta$ as $\mathcal{C}(\theta) := (2N)^{-1}\|g(X) - y\|_2^2$, where the norm denotes the countable $\ell^2$ norm. This is the standard mean squared error loss.
    \item If necessary for convenience, define a \textit{template model} as a pair $(\Theta, X)$ and an \textit{instantiated model} as a triple $(\Theta, X, \theta)$. Each of these objects carries implied values of $N, M$, and $K$. 
    \item We call a template model $(\Theta, X)$ \textit{solvable} if for all $y$ there exists $\theta$ such that $\mathcal{C}(\theta) = 0$. 
\end{enumerate}
\end{definition}

\section{Overparameterized Networks}
\label{sec-overpar-1-0}

Recent results from \cite{cheewa-5, ch-7} have shown that in the overparameterized setting, the dynamics of gradient descent are deformable to linear interpolation in output space, at least away from the `bad regions' in parameter space where the Jacobian matrix of the network outputs with respect to the parameters is not full rank. More precisely, there is a continuous deformation of output space which converts every gradient descent path which does not encounter such a region into a straight line \cite{cheewa-5}. This setting also grants convergence speed estimates. 

Jacobian rank loss presents a problem for the interpretation of gradient descent: continuous or `true' gradient descent considered as a solution to the gradient flow ODE is redirected by rank-loss regions and changes direction unpredictably. However, practical implementations of gradient descent such as the ever-popular forward Euler stochastic gradient descent will almost surely `miss' or `tunnel through' the measure-zero rank-deficient regions. However, this does not mean that Jacobian rank loss is irrelevant. Rather, it implies that practical gradient descent is not necessarily well modeled by the ideal gradient flow at any point after the trajectory has crossed a rank-deficient barrier. 

It has long been known in the literature \cite{ntk}  that in the infinite-width limit of a shallow neural network, the Jacobian is constant through gradient descent. This heuristically suggests that in the large parameter limit the Jacobian is generically always full rank. In this section, we will describe some ways in which this inference may be extended (or not) to the case of large numbers of parameters, i.e. $K, M > N$. This allows us to better understand the qualitative training behavior of networks of arbitrary depth at large (but still finite) width. 

\subsection{Other Work}

To our knowledge, little work has been done on the rank of the output-parameter Jacobian. Some related works are as follows:

\begin{itemize}
    \item Some analysis of the clean behavior of gradient descent was performed in \cite{du}. Their work assumes that `no two inputs are parallel', i.e. that $X$ is full rank the in language of our work, and they work with shallow networks only. We believe that this work contributes towards extending such analysis to the more general case of deep learning, and to more general activation functions.
    
    \item The papers \cite{feng, belrose} also investigate the Jacobians of neural networks as they relate to gradient descent, but they are not the same ones discussed here. Note that the Jacobian discussed here is `$df/d\theta$' not `$df/dx$'.

    \item The relation of the Jacobian [in the sense of this paper] rank and generalization performance is investigated in detail in \cite{oymak}, but the setting differs greatly from this work. 
\end{itemize}

\subsection{Preliminaries}

We will only deal with networks that are strongly overparameterized. We will show shortly that, as expected, strongly overparameterized networks are usually solvable. 

\begin{definition}[Strongly Overparameterized]
    We say that a template model $(\Theta, L, X)$ is \textit{strongly overparameterized} if $M > N$ and $K > N$. Note that the former implies the latter for essentially every neural network model. 
\end{definition}

Intuitively, it is sensible that wider matrices will fail to be full rank less frequently. However, there is an important wrinkle: if $X$ itself is rank deficient, then Jacobian rank deficiency may be more common than expected. The following results (regarding several common feed-forward models) can be summarized as follows:
\begin{enumerate}
    \item If $X$ is full rank then the model is solvable and $D$ is almost always full rank.
    \item If $X$ is not full rank, it does not necessarily follow that $D$ is not full rank. 
\end{enumerate}

We will need to use the following technical lemma. Please observe that in this paper we will use the Einstein summation convention for repeated indices, but we intend to suppress the summation over a particular index if the index is repeated on one side of an equation, but not on the other side. For example, we write the definition of the Hadamard product of vectors as $(u \odot v)^i = u^i v^i$, and no summations are implied. 

\begin{definition}[Broadcast Vectorized Tensor Product]
    Let $A \in \R^{n \times s}, B \in \R^{n \times t}$. Define a matrix $A \bar{\otimes} B \in \R^{n \times st}$ by letting $(A \bar{\otimes} B)^i = \text{Vec}[A^i \otimes B^i]$ (where the tensor product is represented by the Kronecker product). We will refer to the column indices of $A \bar{\otimes} B$ with a `pair index' $(\alpha, \beta)$. The above definition can also be stated as follows:
    \begin{align}
        (A \bar{\otimes} B)_{(\alpha, \beta)}^i = A^i_\alpha B^i_\beta
    \end{align}
    Observe that the operation $\bar{\otimes}$ is commutative up to column permutation.
\end{definition}

\begin{lemma}
\label{lem:rank}
     Let $A \in \R^{n \times s}, B \in \R^{n \times t}$. Assume that $s, t \geq n$ and $B$ has no zero rows. Then if $A$ is full rank, it follows that $A \bar{\otimes} B$ is full rank. 
\begin{proof}
    For brevity, denote $Z = A \bar{\otimes} B$. We proceed by contrapositive.
    Assume that $Z$ is not full rank. Then, since $s, t \geq n$, $st \geq n$, so the rows of $Z$ are linearly dependent. By definition, there must exist $c \in \R^n$ such that $c \neq 0$ and $c_i Z^i = 0$. Therefore there must exist $\hat{\imath}$ such that $c_{\hat{\imath}} \neq 0$. Since $B$ has no zero rows, there must exists $\hat{\beta}$ such that $B^{\hat{\imath}}_{\hat{\beta}} \neq 0$. Define $\eta \in \R^n$ by $\eta_j := c_j B^j_{\hat{\beta}}$. Since $\R$ has no zero divisors, we know that $\eta_{\hat{\imath}} = c_{\hat{\imath}} B^{\hat{\imath}}_{\hat{\beta}} \neq 0$, so $\eta \neq 0$. Now, pick any $\alpha \in [s]$. Since $c_i Z^i = 0$, in particular we have $0 = c_i Z^i_{(\alpha, \hat{\beta})} = c_i B^i_{\hat{\beta}} A^i_\alpha = \eta_i A^i_\alpha$. Since $\alpha$ was arbitrary, it follows that $\eta_i A^i = 0$, so $A$ has linearly dependent rows. Since $s \geq n$, $A$ is not full rank.
\end{proof}
\end{lemma}

\begin{remark}
    It is easy to see that if either $A$ or $B$ has a zero row, then $A \bar{\otimes} B$ is not full rank. It is also the case that if $A^i = aA^j$ and $B^i = bB^j$, then $A \bar{\otimes} B$ has nullity at least 1. However, in general the linear dependence relations of $A$ and $B$ interact unpredictably, and $A \bar{\otimes} B$ is usually, though not always, full rank. 
\end{remark}

\subsection{Linear Model}
\label{sec:linear}

It is most natural to begin by recalling classical underdetermined linear regression. Consider the optimization problem $\min_w \frac{1}{2}\|w\|_2^2$ s.t. $X^Tw-y=0$. By introducing the Lagrangian $L(w, \lambda) = \frac{1}{2}\|w\|_2^2 - \langle \lambda, X^Tw-y\rangle$ and differentiating, we obtain the saddle point conditions $w = X\lambda$ and $X^Tw - y = 0$. Substituting obtains $X^TX\lambda = y$. If we assume that $X^TX$ is invertible, i.e. that $X$ is full rank (recall these are real matrices), then $\lambda = (X^TX)^{-1}y$. Finally, another substitution obtains $w = X(X^TX)^{-1}y$ given by the Moore-Penrose inverse of $X^T$. This process is quite familiar and will be used in the later arguments as a step. 

However, note that $X^T$ is surjective as a linear transformation if and only if it is full rank. Therefore, if $X$ is not full rank, the problem has no solution unless $y$ is in the span of $X^T$, which is measure zero in $\R^N$. Note that the critical step involves the assumption that $X$ is full rank: if the data of $X$ are not linearly generic, we cannot fit arbitrary labels $y$ using a linear model. We next turn to the quite simple answer for the Jacobian rank:

\begin{proposition}
    Let $\Theta$ be linear regression. Then the following are equivalent:
\begin{itemize}
    \item $D$ is full rank.
    \item $X$ is full rank.
    \item $(\Theta, X)$ is solvable. 
\end{itemize}
\begin{proof}
    In the language we have developed for supervised learning, we can express $f_\theta(x) = \langle x, \theta\rangle$, and $g_\theta(X) = X^T\theta$. It follows that $D = X^T$. The first equivalence is then trivial. The second equivalence follows from the above discussion.
\end{proof}
\end{proposition}

These three conditions are not equivalent for nonlinear models. We shall see shortly that in the strongly overparameterized case, the solvability of neural network models follows from the solvability of linear regression. The relationship of the ranks of $X$ and $D$, however, differs depending on the network architecture. 

\subsection{Abstract Nonlinearity}

In the case where our network is composed of a single affine transform followed by some abstract fixed non-linearity, we have essentially the same properties as in the linear case. First, the technicalities:

\begin{definition}[Fixed Nonlinearity Affine Network]
    Let $\sigma : \R^M \lra \R$ be a surjective $C^1$ submersion and $\Theta(\theta)(x) = \sigma \circ A(x)$ where $A(x) = Wx + b$, $\theta = (W, b)$, and $W \in \R^{M\times M}$. We say that $\Theta$ is a \textit{fixed nonlinearity affine network} on $\sigma$. 
\end{definition}

\begin{proposition}
    If $X$ is full rank, then $\Theta$ is solvable. 
\begin{proof}
    Consider any $y \in \R^N$. For any $i \in [n]$, surjectivity of $\sigma$ implies that $\sigma^{-1}(y^i) \subseteq \R^M$ is nonempty. Therefore, we may pick $\tilde{y}^i \in \sigma^{-1}(y^i)$ for each $i$. The $\tilde{y}^i$ assemble into the rows of a matrix $\tilde{Y} \in \R^{N \times M}$. Define $Z = X(X^TX)^{-1}\tilde{Y}$. It follows that $X^TZ = \tilde{Y}$. Let $W = Z^T$ and $b=0$. Then $\sigma(A(X_i)) = \sigma(Z^TX_i) = \sigma(\tilde{Y}^i) = \sigma(\tilde{y}^i) = y^i$. 
\end{proof}
\end{proposition}

\noindent
We can now calculate derivatives of the network with respect to the weights:
\begin{align*}
    \frac{\partial}{\partial W_\beta^\alpha} f_\theta(X_i) = \nabla \sigma |_{A(X_i)} \cdot \frac{\partial}{\partial W_\beta^\alpha}(WX_i + b) = (\partial_j \sigma)|_{A(X_i)}\frac{\partial}{\partial W_\beta^\alpha} [W^j_k X_i^k]\\
    = (\partial_j \sigma)|_{A(X_i)} \delta_{\alpha j} X_i^\beta = (\partial_\alpha \sigma)|_{A(X_i)} X_i^\beta = \text{Vec}[\nabla \sigma|_{A(X_i)} \otimes X_i]_{(\alpha, \beta)}
\end{align*}
The derivatives with respect to the biases are much simpler:
\begin{align*}
    \frac{\partial}{\partial b_\gamma} f_\theta(X_i) = \nabla \sigma |_{A(X_i)} \cdot \frac{\partial}{\partial b_\gamma}(WX_i + b) = (\partial_\gamma \sigma)|_{A(X_i)}
\end{align*}
Define $\nabla \sigma|_{A(X)} \in \R^{N \times M}$ as the matrix whose $i$th row is $\nabla \sigma|_{A(X_i)}$. Then we have the result that 
\begin{align*}
    D = \begin{bmatrix} \nabla \sigma|_{A(X)} \bar{\otimes} X^T & \nabla \sigma|_{A(X)} \end{bmatrix}
\end{align*}
If we work with a slightly modified model that has no local bias, the result is even simpler: $D = \nabla \sigma|_{A(X)} \bar{\otimes} X^T$. We may now turn to making claims about the rank of $D$.

\begin{proposition}
\label{prop:mid-fullfull}
    If $X$ is full rank, then $D$ is full rank. 
\begin{proof}
    Since $\sigma$ is a submersion, $\nabla \sigma \neq 0$ everywhere. In particular, $\nabla \sigma|_{A(X)}$ has no zero rows. The result then follows by the Lemma \ref{lem:rank}.
\end{proof}
\end{proposition}

\noindent
This shows in particular that continuous gradient descent drives the system to the zero-loss minimizer and that numerical gradient descent does not significantly diverge from this behavior.
If $X$ is not full rank, then the situation becomes more complicated. However, we can state two general facts:

\begin{proposition}
Assume that $A(X)$ has no duplicate rows. Then there exists $\sigma$ such that $D$ is full rank. If $A(X)$ also has no zero rows, then $D$ has full rank in the bias-omitted model.
\begin{proof}
    Pick an arbitrary full rank matrix $G \in \R^{N \times M}$. Then by multivariate Hermite interpolation, there exists a polynomial $\sigma$ such that $\nabla \sigma |_{A(X)} = G$. The first result is trivial, the latter follows by Lemma \ref{lem:rank}. 
\end{proof}
\end{proposition}

\noindent
In particular, this shows that rank deficiency of $X$ does not always result in rank deficiency of $D$; i.e. the converse of Proposition \ref{prop:mid-fullfull} is not true. If $X$ is rank deficient, then $\nabla \sigma$ must be complicated enough to compensate for this rank deficiency, or else $D$ will also be rank deficient.

\subsection{Feed-Forward Neural Networks}
\label{sec:ffnn}

Regarding the question of solvability, we will supply the proof of Theorem \ref{thm-overpar-expl-min-1-0}. First, we prove a lemma:

\begin{lemma}
\label{lem:image}
Take any map $\sigma : \R^M \lra \R^M$ which is a local diffeomorphism at at least one point. Then for any dataset $X$, there exists an affine map $A = (W, B)$ and a dataset $Y$ such that $W\sigma(Y_i) + B = X_i$ for all $i$, where $Y$ is a smooth function of $W, B$, and $X$. 
\begin{proof}
    Let $z \in \R^M$ be a point at which $\sigma$ is a local diffeomorphism. It follows that there exists an open box set $R = \prod_{j\in [N]} (a_j, b_j)$ where $a_j < b_j \forall j$ such that $\sigma(z) \in R$ and $\sigma^{-1}|_R$ is a diffeomorphism. Since all nondegenerate open boxes are affine equivalent, there exists an invertible affine transform $A = (W, B)$ such that $X \subseteq WR + B$. Therefore, there exists a unique $Y_i = \sigma^{-1}(A^{-1}(X_i))$ for each $i$. Since $A$ is diffeomorphic and $\sigma$ is a diffeomorphism on this region, $Y$ is a smooth function of $W, B, X$. 
\end{proof}
\end{lemma}

Note that the data matrix, previously referred to as $X$, is here more conveniently referred to as $X^{(0)}$. We continue now with the proof of Theorem \ref{thm-overpar-expl-min-1-0}:

\begin{proof}
To begin with, we consider the linear network obtained from $L=0$, that is, the output layer is the only layer. Then, zero loss minimization of the cost is equivalent to solving
\eqn 
	W_1 X^{(0)} + B_1 = Y_\omega \in \R^{Q\times N} \,.
\eeqn 
This slightly generalizes the problem covered in Section \ref{sec:linear}.
Let us write $W_1 = A (X^{(0)})^T \in\R^{Q\times M}$ where $A\in\R^{Q\times N}$. Then, generically, $X^{(0)}$ has full rank $N<M$, so that $(X^{(0)})^T X^{(0)} \in\R^{N\times N}$ is invertible, and we obtain from
\eqn 
	A (X^{(0)})^T X^{(0)}  = Y_\omega -B_1 \,,
\eeqn 
that 
\eqn\label{eq-W1-sol-linNet-1-0}
	W_1 = (Y_\omega -B_1) ((X^{(0)})^T X^{(0)} )^{-1}(X^{(0)})^T \,.
\eeqn 	
It follows that $W_1 X^{(0)}=Y_\omega -B_1$, and we have found the explicit zero loss minimizer.

Next, we assume $L$ hidden layers. In the output layer, we have $Q\leq M$. 

Assume that $\sigma : \R^M \lra \R^M$ is a local diffeomorphism at at least one point.
Consider that $Y_\omega$ embeds as $S^{(L+1)}$ in $\R^N$ by setting all trailing coordinates to zero. Let $P:\R^M \lra \R^Q$ be the projection onto the first $Q$ coordinates. Then recursively let $W'_j, B'_j$, and $S^{(j-1)}$ be defined by Lemma \ref{lem:image} such that $W'_j \sigma(S^{(j-1)}_k) + B'_j = S^{(j)}_k$ for all $k$, for all $j$ such that $2 \leq j \leq L+1$. Finally, we must solve the linear regression $W_1X^{(0)} + B_1 = S^{(1)}$, which has a closed-form solution for $X^{(0)}$ full rank, i.e. generic data: pick $B_1$ arbitrarily and set
\begin{align}
    W_1 = (S^{(1)} - B_1) ((X^{(0)})^T X^{(0)} )^{-1}(X^{(0)})^T 
\end{align}
Let $W_j, B_j$ = $W'_j, B'_j$ for $2 \leq j \leq L$, but let $W_{L+1} = PW'_{L+1}$ and $B_{L+1} = PB'_{L+1}$. This set of weights and biases is constructed to give exactly zero loss, which proves the claim. 
\end{proof}

It should be clear that the model is extremely redundant: we need to jump through a variety of hoops to pull the data back through the layers, during which many arbitrary choices are made, and in the end we just end up using a single-layer linear regression anyway. This reflects the fact that a deep neural network is massive overkill in the high-width tail, i.e. strongly overparameterized regime. 

Now, we turn to characterizing the rank of $D$. We would like to understand the geometry of the rank-deficient set. Consider the two-layer one-output network $f(x) = w^T \sigma(Wx + B)$, where $\sigma$ is a submersion (such as a mollified ReLU, for example). Then 
\begin{align}
    \frac{\partial}{\partial W^\alpha_\beta} f(x) = \frac{\partial}{\partial W^\alpha_\beta} w_i \sigma^i(Wx+B) = w_i (\partial_\alpha\sigma^i|_{Wx+B}) x^\beta
\end{align}
It follows that
\begin{align}
    D = \begin{bmatrix}
        \sigma(WX + B) &
        w_i \nabla \sigma^i|_{WX + B}  \bar{\otimes} X^T &
        w_i \nabla \sigma^i|_{WX + B}
    \end{bmatrix}
\end{align}
Where the order of the variables is $w, W, B$. Therefore, for $D$ to be full rank (assuming that $X$ is full rank) it is sufficient that $(d\sigma_{Wx_j + B})w \neq 0$ for all $j$. Since $\sigma$ is a submersion, this is equivalent to $w \neq 0$. This set of parameters has codimension $M$, and therefore is very unlikely to be encountered by a gradient flow path since $M$ is large---we leave a detailed quantitative analysis of this to future work. 

It is clear that there are many cases where $D$ may be rank deficient independently of the rank of $X$. For example, if the activation is ReLU, and all the data is mapped into $\R^M_{-}$ in any layer, then $D$ is rank zero. On the other hand, perhaps $\sigma$ is a diffeomorphism, but all the weight matrices and biases are zero. Then $D$ has rank exactly 1, from the output layer bias. In this sense, the extra depth has, strictly speaking, only made the situation worse, since the addition of the redundant layers has also introduced many opportunities for rank loss. However, if we relax the situation slightly, we can apply our earlier result to obtain a guarantee:

\begin{proposition}
    Suppose $\sigma$ is a submersion and $W_2, \dots, W_{L+1}$ are full rank. Then if $X$ is full rank, $D$ is full rank. 
\begin{proof}
    Since $W_j$ is full rank for $j > 1$, $X^{(L+1)}$ is equal to a sequence of submersions applied to $W_1X^{(0)} + B_1$. In other words the network may be written as $\theta(W_1X^{(0)} + B_1)$ for $\theta$ a submersion. The result follows by Proposition \ref{prop:mid-fullfull} (note that the assumption of surjectivity was not necessary in that proof).
\end{proof}
\end{proposition}

This is, for instance, true in an open neighborhood in parameter space around the solution constructed in the proof of Theorem \ref{thm-overpar-expl-min-1-0}, since full rank is an open condition on the space of matrices. We leave a further investigation of the geometry of the rank-deficient set and its consequences for gradient flow to future work.

\section{Proof of Theorem \ref{thm:alt}}
\label{sec:appendix}

\begin{proof}
To begin with, we consider the linear network obtained from $L=0$, that is, the output layer is the only layer. Then, zero loss minimization of the cost is equivalent to solving
\eqn 
	W_1 X^{(0)} + B_1 = Y_\omega \in \R^{Q\times N} \,.
\eeqn 
This slightly generalizes the problem covered in Section \ref{sec:linear}.
Let us write $W_1 = A (X^{(0)})^T \in\R^{Q\times M}$ where $A\in\R^{Q\times N}$. Then, generically, $X^{(0)}$ has full rank $N<M$, so that $(X^{(0)})^T X^{(0)} \in\R^{N\times N}$ is invertible, and we obtain from
\eqn 
	A (X^{(0)})^T X^{(0)}  = Y_\omega -B_1 \,,
\eeqn 
that 
\eqn\label{eq-W1-sol-linNet-1-0}
	W_1 = (Y_\omega -B_1) ((X^{(0)})^T X^{(0)} )^{-1}(X^{(0)})^T \,.
\eeqn 	
It follows that $W_1 X^{(0)}=Y_\omega -B_1$, and we have found the explicit zero loss minimizer.

Next, we assume $L$ hidden layers. In the output layer, we have $Q\leq M$. Then, zero loss minimization requires one to solve
\eqn\label{eq-output-solve-1-0}
	W_{L+1} X^{(L)} = Y_\omega - B_{L+1} 
	\;\;\;\in \R^{M\times N} \,.
\eeqn 
We may choose $W_{L+1}\in\R_+^{Q\times M}$ to have full rank $Q$, so that $W_{L+1}W_{L+1}^T\in\R_+^{Q\times Q}$ is invertible, and determine $b_{L+1}$ so that 
\eqn 
	X^{(L)} = W_{L+1}^T(W_{L+1}W_{L+1}^T)^{-1}( Y_\omega - B_{L+1} )
	\;\;\; \in\R_+^{M\times N} \,.
\eeqn 
For instance, one can choose $b_{L+1}$ to satisfy $-(W_{L+1}W_{L+1}^T)^{-1}b_{L+1}=\lambda u_Q$ where $u_Q=(1,1,\dots,1)^T\in\R^Q$ is parallel to the diagonal. Then, for $\lambda>0$ sufficiently large, all column vectors of $(W_{L+1}W_{L+1}^T)^{-1}Y_\omega$ are translated into the positive sector $R_+^Q$. Application of $W_{L+1}^T$ then maps all resulting vectors into $\R_+^M$ because the components of $W_{L+1}\in\R_+^{Q\times M}$ are non-negative. This construction is similar as in \cite{cheewa-1}.

In particular, we thereby obtain that every column vector of $X^{(L)} \in\R_+^{M\times N}$ is contained in the domain of $\sigma^{-1}:\R_+^M\rightarrow\R^M$. To pass from the layer $L$ to $L-1$, we then find 
\eqn 
	X^{(L-1)} = W_L^{-1}\sigma^{-1}(X^{(L)}) - W_L^{-1} B_L
\eeqn 
where $b_L$ can be chosen to translate all column vectors of $W_L^{-1}\sigma^{-1}(X^{(L)})$ into the positive sector $R_+^M$ along the diagonal, with $-W_L^{-1}b_L=\lambda u_M$ and $\lambda>0$ sufficiently large.

By recursion, we obtain
\eqn\label{eq-Xell-rec-expl-1-0}
	X^{(\ell-1)} = W_\ell^{-1}\sigma^{-1}(X^{(\ell)}) - W_\ell^{-1} B_\ell
	\;\;\;
	\in\R_+^{M_{\ell-1}\times N}
\eeqn
for $\ell=2,\dots,L$, and
\eqn 
	W_1 X^{(0)} = \sigma^{-1}(X^{(1)}) - B_1
\eeqn 
where $X^{(1)}$ is locally a smooth function of $Y_\omega,(W_j,b_j)_{j=2}^{L+1}$. For generic training data, $X^{(0)}\in\R^{M\times N}$ has full rank $N<M$, and in the same manner as in \eqref{eq-W1-sol-linNet-1-0}, we obtain that
\eqn 
	W_1 = (\sigma^{-1}(X^{(1)}))((X^{(0)})^T X^{(0)} )^{-1}(X^{(0)})^T
\eeqn 
where we chose $b_1=0$ without any loss of generality. Hereby, we have constructed an explicit zero loss minimizer.
%Finally, we note that when $L>0$, the case $Q<M$ can be easily mapped to \eqref{eq-output-solve-1-0} by embedding $Y_\omega\in\R^{Q\times N}$ into $\R^{M\times N}$ via the addition of $M-Q$ rows of zeros.

The arbitrariness in the choice of the weights $W_j$ and biases $b_j$, $j=2,\dots,L+1$, implies that the global zero loss minimum of the cost is degenerate.
\end{proof}

$\;$\\
\noindent
{\bf Acknowledgments:} 
T.C. thanks P. Mu\~{n}oz Ewald for discussions.
T.C. gratefully acknowledges support by the NSF through the grant DMS-2009800, and the RTG Grant DMS-1840314 - {\em Analysis of PDE}.   
\\

\end{document}